# Parallel, Modular Fabric Actuators for a 2-DOF Pneumatic Sleeve

Rainier F. Natividad and Raye Yeow Chen-Hua, *Member, IEEE*

*Abstract* — The shoulder is responsible for the movement of the entire upper limb. It is capable of articulating in three degrees-of-freedom (DOF), enabling the arm to perform manipulation actions with incomparable dexterity. Exoskeletons targeting the shoulder must be able to emulate its complex kinematics. While typical architectures have proven to be useful, they employ complex and usually inadequate techniques to match the mobility of the shoulder. A new 2-DOF soft robotic shoulder exoskeleton is presented. Modular soft robotic actuators with separated inflation modules on the exoskeleton are able to emulate the humerus' natural movement. These actuators were organized into two antagonistic pairs and operate in a parallel configuration. We present the design and functionality of the actuators powering the exoskeleton. The actuator design enables it to perform 3-D bending with insignificant resistance. We measured its performance through a series of static and dynamic tests. Depending on the configuration, the actuator can generate up to a maximum torque of 15.54 N-m and can respond with a minimum rise time of ~2s when excited by a step input. We demonstrated its ability to perform assistance of shoulder movements. It is capable of reaching any point on the humerus' workspace from any starting position.

## I. Introduction

The human arm is responsible for a majority of movements necessary for activities of daily living (ADL) [1]. As a result, humans are extremely reliant on their arms in order to live an optimal life. The humerus is the most mobile bone in the human body due to its ability to effectively and efficiently steer the human arm to perform its desired tasks [2]. It is connected to the thorax through the shoulder complex and is, thus, considered as the structural base of the entire human arm. The mobility of the humerus is due to the unique musculoskeletal structure of the shoulder complex. While it externally shows a single joint, its skeletal structure is composed of four joints: the scapulothoracic joint, the sternoclavicular joint, the acromioclavicular joint, and the glenohumeral (GH) joint. These joints connect the sternum, clavicle, scapula and humerus to form the shoulder complex. Although the GH joint is a ball-and-socket type of joint that is primarily responsible for movement [3], the 4 joints must work in concert in order to effectively move the humerus. Any amount of movement of the shoulder complex necessitates the coordinated motion of all joints. This synchronized motion is called the scapulohumeral rhythm. Failure to adhere to this rhythm will result in the inefficient operation of the arm and an eventual injury[4].

Exoskeletons catering to the shoulder must, therefore, account for the scapulohumeral rhythm. Clinical upper-limb exoskeletons are typically used to treat disorders that inhibit shoulder movement [5], [6] and, consequently, hinder a person's ability to perform ADLs. Typically, these exoskeletons are constructed using techniques pioneered in industrial robotics. Rigid electric motors are placed concentrically with the humeral head while rigid links are attached parallel to the humerus, connecting the motors to the arm. In order to account for the scapulohumeral rhythm, these exoskeletons must translate the position of the motors during shoulder movement[5]. Alternatively, designers also employ an additional link in order to account for translation of the humeral head. However, the addition of such a link increases the likelihood of joint misalignment[6]. Moreover, these designs weigh substantially relative to the weight of a human, which adds unwanted inertia to the arm and restricts the mobility of the user.

In contrast to traditional robotic designs, soft robotics utilizes naturally flexible materials both as prime movers and structural links. This archetype of robotics behaves similar to continuum structures [7] as compared to a system of rigid bodies that typically characterize traditional robots. The distinct compliance and usage of non-rigid power transmission elements have enabled soft robots to more closely emulate the movement of animals. They are uniquely suited to power robotic exoskeletons due to their compliance, which results in the ability to conform to the external structure of the human body, and accurately follow the movement of the shoulder complex. Soft robots can be created in various forms, with each form having its own method of power transmission, such as cable-driven devices[8], shape-memory actuators[9], combustion[10], magnetism[11], and pressurized fluids[12]. Cable-driven exoskeletons[13] most closely mimic the structure of the muscular system but require accurate positioning of anchor points. Meanwhile, pressurized fluids, specifically compressed air, have proven to be a popular choice for driving shoulder exoskeletons. However, most fluidic exoskeletons are unable to provide sufficient mobility[14]–[16]. O'Neill et al constructed a 2-DOF soft exoskeleton by performing sequential abduction and horizontal flexion/extension motions using two groups of actuators in a serial configuration. They mounted a single abduction actuator on the medial arm and the lateral torso which stiffens upon pressurization and attached 2 rotational actuators which rotate the abduction actuator to create the 2$^{nd}$ DOF [17]. While their exoskeleton also possesses 2-DOF, it is unclear if it is capable of performing reaching actions, which is essential in performing a variety of ADLs [18]. Moreover, their exoskeleton was only able to provide support to up to 55.1° in abduction, and 56.7° of horizontal flexion. Soft robotic technology has also been applied to other limbs such as the leg[19] and more popularly, the hand[20].

We present in this study a soft robotic, pneumatic, shoulder exoskeleton (Fig. 1) with 2-DOF that can generate assistive torque to any position in the humerus' workspace starting from

Research supported by MOE AcRF Tier 2 (R-397-000-281-112)

Rainier F. Natividad and Raye Yeow Chen-Hua are with the Department of Biomedical Engineering at the National University of Singapore (NUS), Singapore (e-mail: *rfnatividad@u.nus.edu*, bieych@nus.edu.sg).

Rainier F. Natividad and Raye Yeow Chen-Hua are also with the Advanced Robotics Center (ARC) at NUS.

Raye Yeow Chen-Hua is also with the Singapore Institute of Neurotechnology (SINAPSE) at NUS

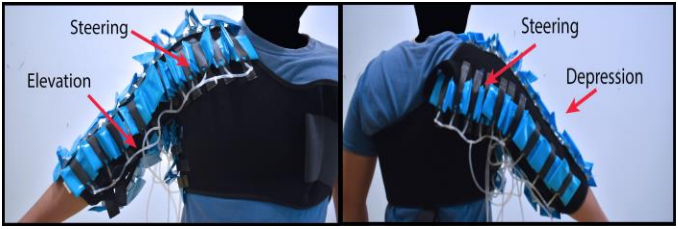

Fig 1: Front (Left) and back (Right) views of the 2-DOF shoulder exoskeleton. The exoskeleton is powered by 2 pairs of anagonistic, modular actuators. The elevation/depression pair and the steering pair are attached perpendicularly.

any position. The exoskeleton is capable of performing forward flexion and extension, abduction and adduction, and horizontal flexion and extension, as well as driving the upper arm to any pose in between these standard positions. At the core of the exoskeleton is a set of modular, pneumatic bending actuators with separated and replaceable inflation modules based in [21], [22]. The actuators generate torque upon the injection of pressurized air. The actuators achieve modularity through the use of two primary components (Fig. 2-A), namely the inflatable modules and the spine to which they are attached. The modular feature of the actuator allows for the creation of distinct regions on each actuator – areas around the joint are designed to bend, while the areas along the torso and upper arm are designed to resist bending without generating torque. Two antagonistic actuator pairs are placed along the length of the humerus. Unlike the majority of upper-limb exoskeletons which feature a serial actuator configuration [17], [23], these actuators are attached in a parallel configuration wherein the actuators directly impart torque onto one side of the upper arm. The synergistic operation of the antagonistic pairs, coupled with the additional flexibility offered by separated modules, give the exoskeleton added mobility. We present a detailed description of the design and operating principle of the actuator, as well as a summary of the fabrication techniques utilized. Its performance was evaluated both statically and dynamically. The various poses of the actuator as a function of the input pressure, as well as its blocked torque was measured while configured in various positions and pressures. Its step response was also measured. Finally, we demonstrate its ability to provide assisting torque to a healthy user.

## II. Design and Operation

The exoskeleton is powered by four actuators. The primary purpose of the actuators is to provide a combination of torques to the upper arm such that the exoskeleton can push the arm to any point on the anatomical workspace in one single motion. In order to do so, the actuators must be able to reach every point regardless of the starting position. Moreover, any exoskeleton, whether hard or soft, must not cause a significant increase in inertia and its center of rotation must always coincide with that of the humeral head. This entails that the midpoint of the actuators in the exoskeleton of the current study must coincide with the humeral head. In order to address these functional requirements, the actuator is composed of a fabric spine and fabric inflation modules. The use of pneumatics reduces the mass of the actuators as compared to hydraulics. Flexible 3-D printed structures are placed on the spine and modules which serve as mating and locking mechanisms (Fig. 2-A). The actuator is activated through the injection of compressed air. The maximum continuous applied pressure was limited to 80kPa, which helps minimize safety risks. The resulting spatial interference between adjacent, inflated modules generates a bending moment that causes the actuator to generate torque. The torque generated by pneumatic actuators is directly related to their size[21], [22]. However, module sizing was restricted to ensure a moderate exoskeleton size that is wearable. Other mechanical properties can also be modified by the size of the modules.

### A. Fabrication

The spine and modules are primarily composed of a Nylon sheet coated with thermoplastic polyurethane (TPU) on both sides. Nylon is the primary structural component of the parts while TPU facilitates fabrication. Fabrication begins by directly generating 3-D printed structures onto the sheets (Fig. 2-B). The sheets are mounted on a fused deposition modelling (FDM) 3-D printer (Lulzbot, Taz 6). Double-sided adhesive tape is placed on the perimeter of the sheets. The print bed must be heated to a minimum temperature of 60°C in order to ensure proper adhesion of the tape. The height of the nozzle is set to 0.50mm above the surface of the Nylon sheet. The structures are then printed normally using flexible TPU-based filament (Polyflex, Polymaker). The use of TPU-based filament is necessary in order to prevent delamination of the structures from the sheets during operation since the TPU filament bonds exceptionally well to the TPU coat of the sheet. The sheets are then cut to size.

At this stage, the spines are complete and are ready for use. Meanwhile, the modules are further processed based on the technique established in [15]. A hole is drilled on the under-side of the module that connects the fluid path from the 3-D structures onto the modules. The sides are then heat-sealed, with the structure positioned inside as shown in Fig. 2-C. Paper-based tape is used to cover the sealing area in order to prevent the TPU on the external side from sticking to the sealer. The module is then flipped inside-out; a third seal is placed on the top. When deflated, the modules resemble a rectangle with a single seam on the top (Fig 2-D).

### B. Modular Hybrid Fabric-Plastic Actuator

The actuator is assembled by combining appropriate modules of varying lengths and widths based on mechanical requirements. A long groove is located on the located on the 3D-printed spine structure, which mates with corresponding beams on the module structure. This pair acts as a guide so that the two components are properly placed. Slots are placed on the spine structure which subsequently mate with studs on the module. The studs snap into place and lock the module into position (Fig. 2-A). A strip of industrial-grade hook fastener is placed on the spine. The spine is subsequently attached to a neoprene sheet (i.e. exoskeleton base). The modules are then attached, and pneumatic lines are connected to each module using polyethylene (PET) connectors. The pneumatic lines start from a pressure regulator or valve and branch out onto the individual modules. Fabric straps are placed along the center of the modules in order to prevent unnecessary inflation along the covered area (Fig. 2-E). Without the straps, the middle section (i.e. section covered by the straps) will inflate but will not contribute in the generation of torque and will decrease actuation speed[7]. In its neutral deflated state, the adjacent modules fold up on the sides. The actuators behave as continuum structures when pressurized. The folded-up modules inflate and spatially interfere with adjacent modules. When unloaded and unattached to external structures, the actuator curls into a smooth spiral shape until each module is tangential (i.e. barely in contact with adjacent modules) to its adjacent module (Fig. 2-F). The separated structure of the inflation modules, as well as the fabric spine and the absence of side seams, grants unprecedented 3-D flexibility to the actuator (Fig. 2-G). This enables the actuator to perform tight 3-D bends while still maintaining its torque output in its active axis of rotation. The contact area (Fig. 2-H), along with the applied pressure, between adjacent modules is primarily responsible for the magnitude of the torque generated [21], [22].

## III. ACTUATOR PERFORMANCE

The modularity of the actuators, along with the intertwined nature of a soft robot's structure and mechanical characteristics suggest that the actuator's performance can be altered by changing its module configuration. In order to gain an understanding of their behaviors, three actuator variants were constructed, with each variant having a total of eight modules installed. The variants differ in the size and pattern of the modules installed. Their geometric parameters are listed in Table 1. Module widths were based on the lower $5^{th}$ percentile of female upper arm diameters for B, and the lower $5^{th}$ percentile of male upper arm diameters for C [24]. Module spacing (d) was set at 25mm in order to minimize the discontinuity in curvature profile of the actuator. While tighter spacing is desired, the size of the 3-D printed locking structures presents a practical minimum. Alternate module patterning (i.e. AAAA, ABAB, ACAC) was chosen since a homogenous pattern induced buckling in preliminary tests. Module A was created as a bridging module for this purpose. The module lengths were then set to 65mm and 90mm to ensure sufficient spatial interference between adjacent modules.

A customized, electronic pressure regulation system was paired to the utilized measurement platforms for the automated execution of the experiments. A Teensy 3.6 (PJRC.com, LLC) supplied control signals to the platforms. A digital potentiometer (MCP41010, Microchip Technology Inc.) supplied DC voltage signals to the pressure regulator (ITV1031, SMC Corp). The output of the regulator was connected directly into the samples. An industrial compressor supplied compressed

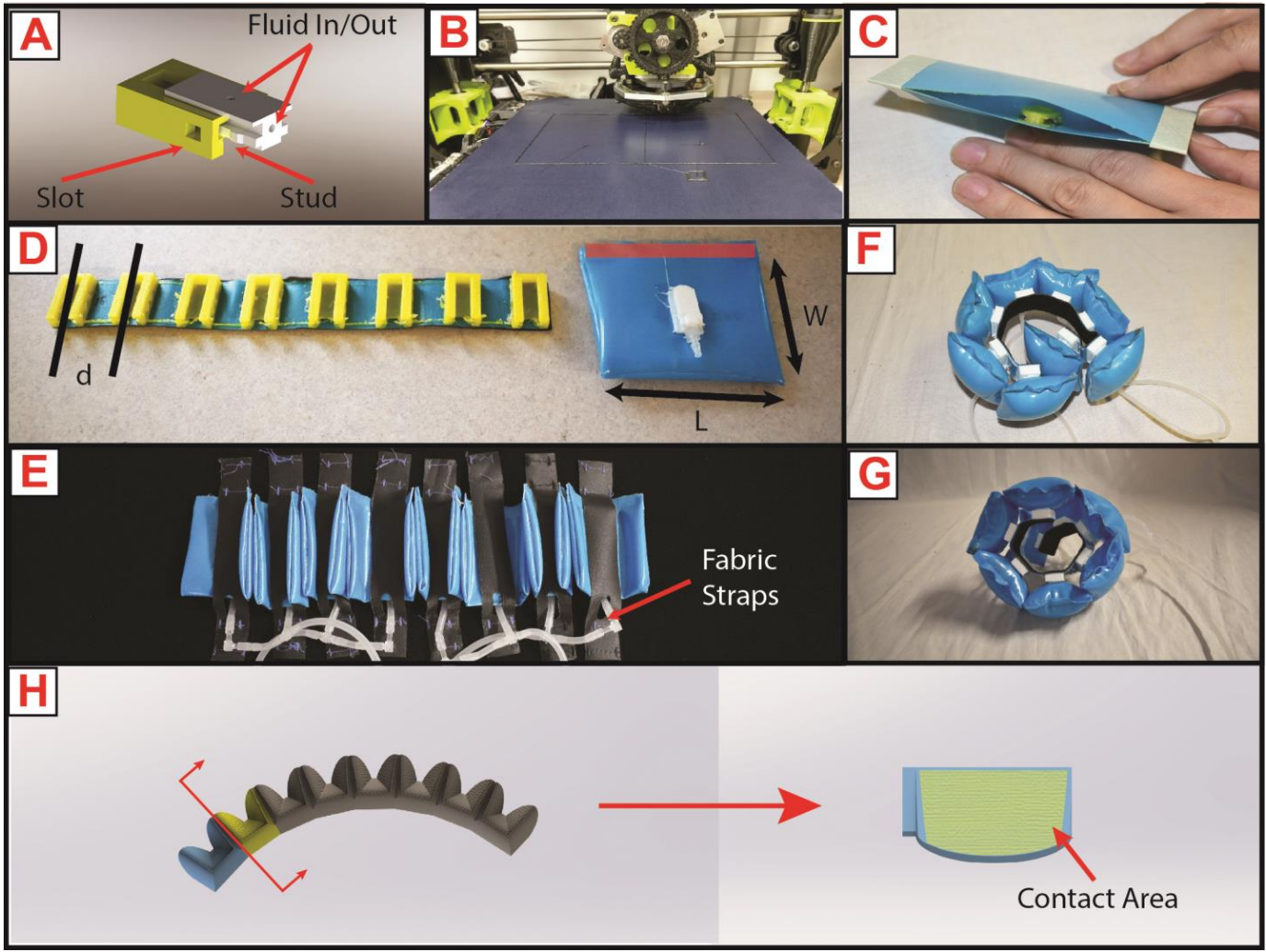

Fig 2: (A) A CAD representation of the 3-D printed structures. (B) 3-D structures are printed directly onto fabric. (C) The modules are fabricated through heat sealing. After sealing the sides, the module is flipped inside-out. D) A completed spine with a completed module. The yellow and white, 3-D printed structures are clearly seen on the spine and module respectively. The red band on the module indicates the location of the heat-seal seam. (E) The assembled actuator. The spine is attached to neoprene through industrial grade hook-and-loop fasteners. Fabric straps are placed across to limit unwanted inflation at the center. (F) The actuator curls during inflation. The strap and neoprene are removed for clarity. At this pose, all the modules are tagential to adjacent modules. (G) The actuator curls into a helix through the application of external forces. Its ability to perform 3-D bending is clearly demonstrated. (H) A schematic representation of a partially curled actuator. A section view of 2 actuators in contact are shown. The yellow shaded area shows the contact area of adjacent modules, generated through spatial interference.

TABLE 1. MODULE DIMENSIONS AND PATTERNING

| Module Variants | Module Length (L) | Module Width (W) |
|---|---|---|
| A | 65mm | 55mm |
| B | 90mm | 55mm |
| C | 90mm | 65mm |
| **Actuator Variant** | **Module Patterning[a]** | |
| D1 | AAAAAAAA | |
| D2 | ABABABAB | |
| D3 | ACACACAC | |

a. The installation arrangement of modules on the variants

air into the regulators at 250kPa. The flow rate into the control system was limited to 60 Standard Liters/min by an FMA-A2317 (Omega Engineering Inc.) mass flow sensor in order to facilitate the comparison of the dynamic performance with other systems. The system has a bandwidth of 10 Hz. The system was configured to have a pressure range of 10-150kPa, with a resolution of 1kPa. Alternatively, it can be vented to the atmosphere effectively setting system pressure to 0kPa.

*A. Experimental Platforms*

Torque measurements were acquired on a platform that is capable of measuring the torque output along two axes. While the actuator is only capable of generating torque along its primary axis, it is important to quantify how much *straightening* torque the actuators can generate when forced into a 3-D pose. The A-A' axis coincides with the actuator's primary axis of bending, while the B-B' axis is perpendicular to A-A' axis. The actuators were placed across 2 arms – a fixed arm and a movable arm. The movable arm was placed on a swing plate, which was responsible for rotating the arm along the A-A' axis. The movable arm was mounted through a pin placed on the swing plate, which enabled rotation along the B-B' axis (Fig. 3-A & Fig 3-B); this pin served as the center of rotation of the B-B' angle. Threaded indexing holes on the swing arm indexer and the swing plate locked the position of the swing plate and the movable arm, respectively. The movable arm was bolted onto the swing plate to set its orientation. By rotating the swing plate with respect to the indexer, and the movable arm with respect to the swing plate, the A-A' and B-B' angles, respectively, were set. A 0° A-A' angle corresponds to a fully folded actuator while 180° corresponds to a straight actuator. The torque measurement setup was meant to simulate the 2-DOF of the human shoulder – abduction/angle of elevation (AoE) and horizontal flexion/angle of plane of elevation (PoE)[25]. Four load cells (FX1901, TE Connectivity) were mounted and placed in custom-designed holders, between the swing plate and the movable arm; and their outputs were translated to torque by factoring in their distance from the center of rotation. In order to measure torque along the A-A' axis, 2 load cells were mounted parallel to the swing plate (Fig. 3-A). The remaining 2 were mounted perpendicularly and were designed to measure torque along the B-B' axis (Fig 3-B). Fig. 3-C shows an actual photo of the platform during an experiment. In the photo, the B-B' angle is 30°.

A separate platform was used to quantify the actuator's step input response and static free bending output. A plastic, rigid bar was attached to the actuator (Fig. 4-A). An 800g weight was attached on the distal end in order to apply resistive torque to the actuator, while 2 IMUS (BN055, Bosch Inc.) were attached to the actuator—one on the posterior end and another on the base of the rigid bar. Measuring the relative position of these IMUs allowed us to determine the bending angle at any given moment. The value of 800g was chosen for the weight since it provides ~1.4N-m of resistance at 90° and also corresponds to D1's maximum torque value at 90° and 80kPa when combined with the length of the rigid bar. The weight was removed during the free bending experiments. The actuators were then hung vertically (Fig. 4-B). Static measurements were acquired by pressurizing reading measurements 30s after pressurization in order to ensure steady-state conditions were reached. A 60s period square wave, with peak-to-peak values of 0-80kPa, was used to perform the dynamic experiment and to ensure that sufficient time was provided for the actuators to reach steady state, where each wave starts with a pressurization input. Experiments were repeated 3 times and the samples were dismounted and remounted before each repetition. The results of the experiments are shown in Fig. 4.

*B. Static Behavior*

Each actuator was able to achieve full bending (>360°) when excited with a pressure of 10kPa when unloaded. Any increase in pressure did not have any effect on its pose. Static conditions where only achieved when each module was tangential to each adjacent module (i.e. adjacent are in point-contact).

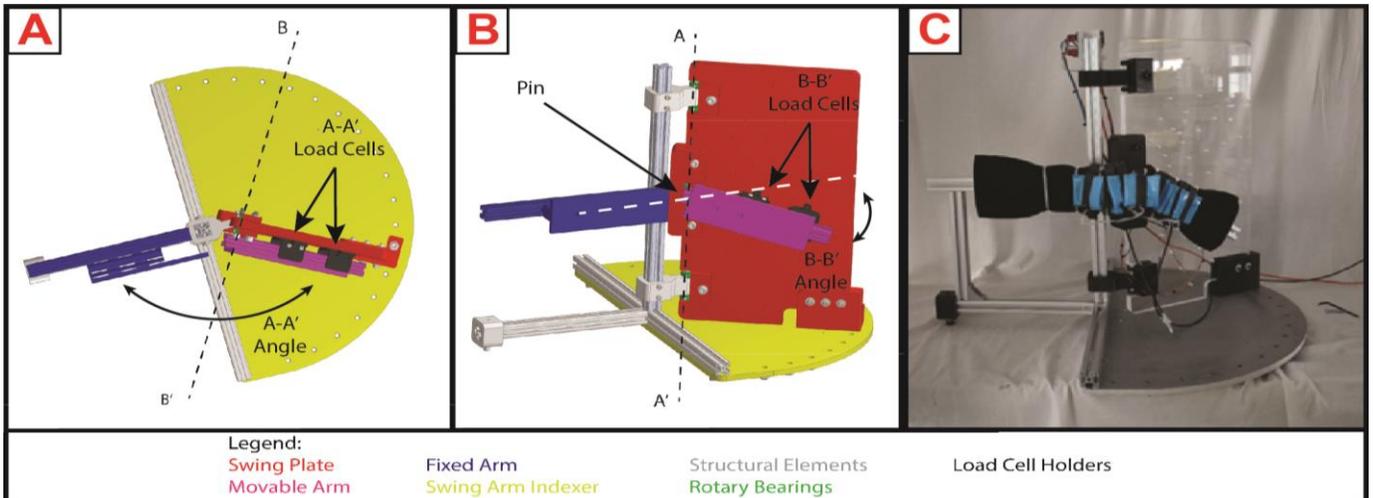

Fig 3: (A-B) A schematic representation of the static blocked torque measurement platform. The parts are color-coded for easy identification. The rotational axes, A-A' and B'B are shown. Positive angular rotations are shown. (C) A photo of the measurement platform.

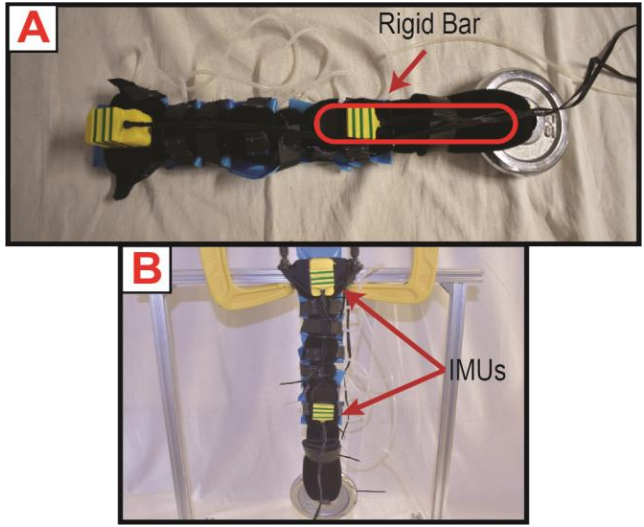

Figure 4: (A) A plastic rigid bar (red outline) and a 800g weight is attached to the actuator for step response measurements. (B) For step response and static free response experiments, the actuators are hung vertically.

Meanwhile, the torque output of the actuator at any input pressure was driven by the contact area the of adjacent modules. A consequence of this relationship is that the maximum torque output at any position would then be dependent on the available contact area. Fig. 4-A shows the results of each variant's torque output as a function of A-A' angle at 30° increments. The measurements were acquired by setting the B-B' angle to 0° and pressurizing the actuator to 80kPa. The actuator exerted maximum torque at 0° and gradually tapered off as angle A-A' increased. Most notably, the rate of decrease tapered off beginning at an A-A' angle of 180°. Effectively, from 180° to 270°, the actuator's torque output became constant. This behavior is due to the fact that from 0° to 180°, the torque generated is a combination of interference forces generated along a bigger surface area since, at this position, the actuators are essentially folded onto themselves. Meanwhile, the unfolding nature of the modules ensure that the contact area is constant when the actuator is positioned between 180° to 270°. The torque output at 80kPa pressure ($T_P$) can be reliably predicted by the A-A' angle (A) through (1) with no apparent overfitting. (1) is a numerical, exponential curve fit with parameters a, b, c, and d. The curves have a minimum coefficient of determination ($R^2$) value of 0.977.

$$T_P = a*exp(b*A) + c*exp(d*A) \quad (1)$$

Fig. 4-B shows the output of the actuator as a function of the input pressure. These measurements were taken at an A-A' angle of 90° and a B-B' angle of 0°. The pressure inside the actuators was varied from 0-80kPa. The output has near-perfect linear correlation between torque and pressure. The data sets of all 3 variants were modeled using (2), with a minimum $R^2$ value of 0.991. Similarly, (2) can be used to predict torque at any given A-A' angle ($T_A$) while f and g are the parameters for (2); g is typically equivalent to zero since the torque when at pressurized is also zero. This behavior is consistent with the behavior of pneumatic bending actuators[21], [22].

$$T_A = f*P + g \quad (2)$$

An increased reduction in available contact area occurs when the A-A' angle and B-B' angle change simultaneously. The effects of this behavior on variant D2 are shown in Fig. 4-C. In this measurement, the A-A' angle was set to 90°; and the B-B' angle was varied from 45° to 0° at 15° increments. This range of motion corresponds to that required of the majority of ADL tasks [18]. The actuator was then pressurized to 80kPa. An overall decrease in the torque output along the A-A' axis was found, as the B-B' angle was increased from 0° to 45°. However, the platform was not able to detect any significant torque output (>0.5N-m) along the B-B' axis. This can be attributed to the ability of each individual module to translate in 3-D as well as the aspect ratio of the modules; they show minimal surface area along the B-B' axis. Effectively, the modules reposition themselves such that there is minimal resistance during operation, resulting in small B-B' torques.

*C. Dynamic Response to Step Input*

Fig. 4-D shows the bending angle response of a weighted actuator; the curves presented have been averaged across the trials. At 0s, the square wave was fed into the actuator. The overall response of the actuators resembled a similar square wave, with an expected time delay. The presence of oscillatory movement was visible in the waveform. There was also variation in the behavior of the actuator during the course of the experiment, primarily attributed to the industrial compressor engaging its motor due to low pressure in the compressor's tank. While this affects the inflow into the pressure regulation system, the overall effects of these variations were minimized by taking multiple measurements. Fig. 4-E and Fig. 4-F show the averaged inflation and deflation responses, respectively. Variants D1, D2, and D3 have inflation rise times of 4.72s, 2.12s and 3.62s, respectively, and deflation rise times of 3.40s, 4.42s and 1.82s, respectively. While previous designs feature longer deflation times, the current actuator has no discernable difference between deflation and inflation times [21]. This change can be attributed to the individualized pneumatic pathing that was utilized. Moreover, no discernable correlation was found between module sizing and response times while the step response resembles that of a 1st-order system with no overshoots.

## IV. THE 2-DOF SHOULDER SLEEVE

The majority of the actuator is composed of nylon fabric, which allows for negligible bending resistance as evidenced by its ability to trace its full range of motion with minimal input. This feature, along with its soft robotic nature, ensures that the shoulder sleeve will have maximum mechanical transparency. These results also entail that the exoskeleton will be able to provide torque assistance regardless of the position of the actuators or the pressure supplied. Nevertheless, the positions of the actuators influence their torque outputs. The three variants effectively exhibit torques of 0.84 N-m, 1.54 N-m, and 1.80N-m at 180°-270°. Load bearing actuators must operate from the 0°-90° range in order to maximize their utility. The actuators are able to apply maximum torque from 0°-90°, with variants D1, D2 and D3 exhibiting peak torques of 10.24 N-m, 11.15 N-m and 15.54 N-m, respectively Meanwhile, they exhibit torques of 1.27 N-m, 4.44 N-m, and 4.66 N-m at 90°, the angle at which the arm imposes maximum static load on the actuator. These actuators would supply approximately 7%, 24.6% and 25.8% of the torque necessary to maintain arm elevation of 90° for a typical, stretched, male arm with a mass of 3.5kg[26]. Nevertheless, users can still benefit from the high peak torques at low elevation angles (i.e. 0°-90°) when performing high velocity, dynamic movements. Moreover, proper use of these actuators, therefore, requires control systems that are able to predict their torque output based on their operating conditions. Equations (1) and (2) can be leveraged for this purpose since they are able to reliably model the actuator's behavior with no overfitting. (1) and (2) can be combined to predict the Torque (T) at any given angle (A) or pressure (P). The actuator must first be characterized in order to obtain its fit parameters for equation (1). This should ideally be done at the maximum pressure of 80kPa. Since its behavior, with respect to varying pressures, is linear, its torque output can be predicted by (3) by obtaining the ratio between 80kPa and the desired pressure(P).

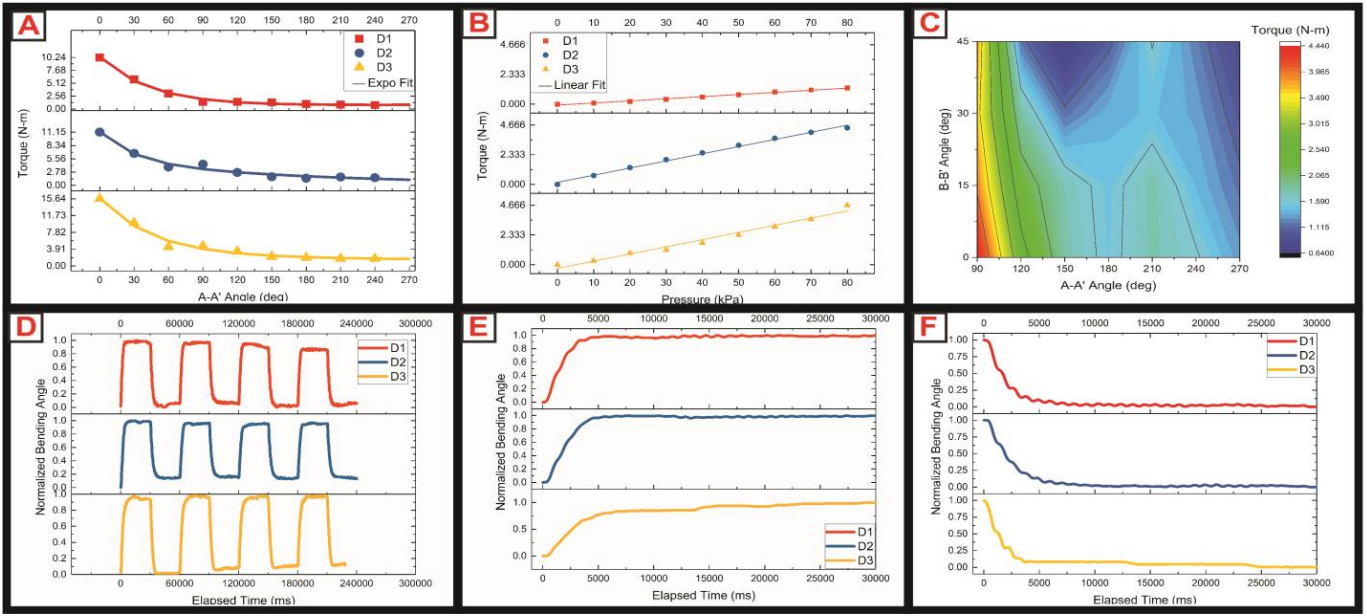

Fig 5: (A) Torque output as a function of A-A' angle when pressurized to 80kPa. (B) Torque output at various inflation pressures, at an A-A' angle set to 90°. Solid lines represent their respective curve fits. (C) A contour plot of torque output at various A-A'angles and B-B' angles for the D2 variant. (D) The waveform in response to a 60s square wave, with 0-80kPa peak-to-peak amplitudes. (E) The averaged step response during inflation. (F) The averaged step response during deflation.

$$T=80P/(a*exp(b*A) + c*exp(d*A)) \qquad (3)$$

Meanwhile, the first order behavior of the actuator suggests that it is feasible for use in exoskeletons; higher-order behaviors such as overshoot are unnecessary risks to users since they might apply massive forces or force the arm to a position outside its safe range. Some oscillation in the behavior of the actuator was present during the course of the experiments, which was expected and can be attributed to the actuator's tendency to buckle when in a free state. Such a problem is not expected during operation of the shoulder sleeve since the actuators are restrained from buckling.

The final design of the exoskeleton is intended to capitalize on the capabilities of the modular actuators. The base of the exoskeleton is a customized neoprene sleeve which acts as a hoop to which the spines can be attached. A loop strap is placed along the torso to ensure that the sleeve remains tight during operation. Additional adjustable straps are also placed on the lateral side of torso on the opposite end. Four actuators, configured as two antagonistic pairs, are placed on the shoulder, with each actuator positioned 90° from each other. Additional modules, positioned as to have minimum available contact area, are added along the upper arm. These modules have minute overlap and are essentially tangential when pressurized. They act as a means to distribute the load generated by the bending actuators throughout the entire upper arm (Fig. 5). These tangential modules bend to a somewhat faint degree, but the amount of curvature is negligible and they effectively act as straight beams. While these are not ideal, they nevertheless function effectively and provide a soft and continuous platform to transfer the torque generated by the actuators onto the whole upper arm. Tangential modules, as opposed to a stopper-like structure, were chosen in order to create a seamless system wherein essentially only inflation modules are attached to the body. A rigid stopper would be extremely uncomfortable for the user, while a soft-bodied stopper would deflect excessively upon inflation. Moreover, such a seamless system could easily be extended to other joints in the arm, such as the elbow and the wrist. However, a system for the entire arm is beyond the scope of this paper.

Using hook-and-loop fasteners, the actuators can be specifically placed on the body of each user. The depression actuator starts from the base of the neck, tracing the superior side of the humerus. The elevation actuator is placed on the inferior side, starting from the lateral area of the ribcage and extending through the armpit and the arm. Meanwhile, the steering actuators are placed on the anterior and posterior arm starting from the sternum and the spine, respectively. The positioning of the actuators takes advantage of their torque-angle curve. Using this configuration, the majority of the load are to be handled by the elevation actuator. This actuator primarily operates in the 0°-90° region, which consequently is the region of maximum torque. Meanwhile, the other actuators reside in the 180°-270° region where the torque-angle relationship is effectively constant. Each antagonistic pair is responsible for each degree of freedom—one pair performs humerus elevation and depression, while another performs rotation of the humerus along the plane of elevation[25]. Moreover, the actuators are mounted in a parallel configuration. This allows the exoskeleton to distribute its force application throughout the entire surface area of the arm. The use of antagonistic actuators also negates the effect of drift during deflation since the agonist will be primarily responsible for *deflating* the actuator. An additional feature of the modular spine is that the torque output of both pairs can be combined such that they can trace trajectories outside a single actuator's axis of rotation. This can be achieved through the simultaneous activation of both pairs. As a result, an exoskeleton worn by a user has a completely spherical theoretical workspace, centered on the humeral head. Realistically, the workspace is limited by the user's maximum range of motion.

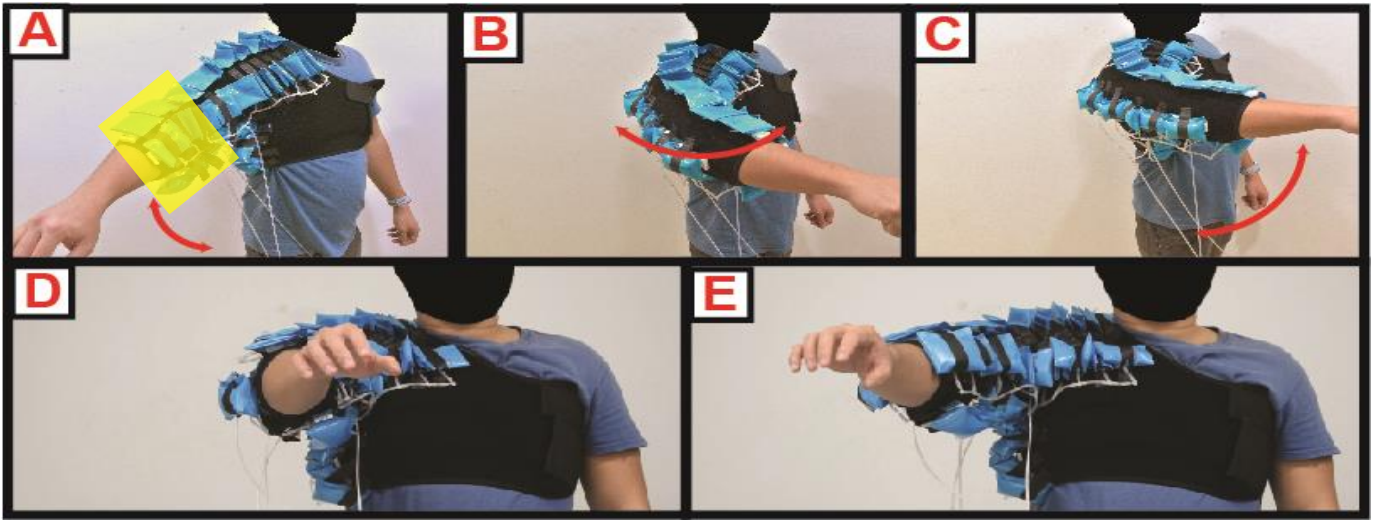

Fig 6: The range of motion of the actuator is shown for an exoskeleton with attached D2 variants. (A) Abduction or adduction. The yellow area highlights non-bending modules that lengthen the moment arm of the bending modules. (B) Rotation along the plane of elevation. (C) Forward flexion or extension. Alternatively, a combination of these movements can be performed through the combined activation of actuators. (D-F) The exoskeleton performing reaching actions. The positions shown were achieved in a single motion, starting from the neutral position.

The operation of the actuators allows the humerus to trace a natural trajectory. From the neutral position, the humerus can perform shoulder abduction and adduction by activating the elevation or depression actuators (Fig. 5-A). The shoulder can also be rotated along the plane of elevation by sequentially activating the elevation and steering actuators (Fig. 5-B). Simultaneous activation of the elevation and steering actuators also grants the ability to perform shoulder flexion or extension (Fig. 5-C). A video of these movements is made available along with this paper. In addition to performing the basic anatomical movements, the controlled activation of both steering actuators and the elevation actuators allows the performance of reaching actions (Fig 5-D, Fig 5-E). Reaching movements allow the exoskeleton to assist the users in a wide variety of ADLs in a seamless manner. While these ADLs may possibly be accomplished using sequential humeral elevation and rotation of the plane of elevation, this series of motions is undoubtedly unnatural. Moreover, the exoskeleton was able to provide support throughout the entire range of motion of the shoulder and this mainly attributed to the minimal mechanical resistance of the actuator coupled with the parallel actuation configuration of the exoskeleton.

## V. HEALTHY SUBJECT TEST

In order to ascertain the exoskeleton's ability to function as a rehabilitative or assistive device for the human shoulder, a healthy, male subject (BW=71kg) was recruited. In accordance with the experiment protocol approved by the NUS Institutional Review Board (N-17-103), the participant was tasked to perform 3 arm motions: abduction and adduction, horizontal flexion and extension, and forward flexion and extension. The subject's informed consent was acquired before the beginning of the test. Wireless EMG sensors (Delsys, Trigno Wireless) were attached to the pertinent muscles: the anterior, lateral and posterior deltoid, the pectoralis major, and the infraspinatus. In order track the subject's movement, one IMU (Bosch, BN055) was placed on the chest and another one was placed on the medial side of the upper arm. The subject then donned the exoskeleton. The EMGs and IMUs were sampled at 2000Hz and 100Hz respectively. At the beginning of the experiment, maximum voluntary contraction (MVC) levels were acquired. Afterwards, the subject was asked to perform arm motions under two conditions: with an unpowered exoskeleton and a powered exoskeleton. Each motion was repeated three times.

IMU data was used to verify if the motions were performed correctly and was subsequently used to separate loading and unloading motions. The separated datasets were then averaged. EMG data was rectified, passed through a 20Hz infinite-impulse response filter with an 80dB stopband attenuation, and normalized with respect to MVC. Its envelope was then acquired by calculating the root-mean square (RMS) envelopes of with a moving window length of 500 samples. The RMS values of each averaged and pre-processed EMG sets were then calculated. Table 2 shows the relative change in EMG RMS for the relevant target muscles across all movements. Overall, the data show that there is a reduction in muscle activation when wearing a powered exoskeleton.

## VI. Conclusion

The exoskeleton presented in this study possesses the useful ability to elevate the arm to any position in a single motion. This emphasizes the capability of this particular exoskeleton to perform reaching motions. Such a device would be a useful tool for performing upper limb, physical rehabilitation, or to function as an assistive device for the shoulder. It utilizes a fabric, modular, pneumatic bending actuator equipped with separated inflation modules. Two antagonistic pairs are mounted in a parallel configuration through the entire circumference of the arm. Each module is supplied with compressed air through branching pneumatic lines, which allows for simultaneous inflation. 3-D bending can be achieved due to the separated structure of the inflation modules. Torque is generated through the spatial interference of modules. The actuator possesses a linear correlation between its torque output and pneumatic pressure and an exponential relationship with its bending angle. While the torque output diminishes from 0° to 180°, it produces constant torque values beyond this range; steering actuators were configured to operate between 180° to 270°.

The actuator behaves as a 1st-order control system when excited by a step input. This suggests that a simple control system can be implemented for steering actuators. Moreover, a test conducted on a healthy subject shows that the current exoskeleton is highly capable of aiding the shoulder. The exoskeleton was able to provide support throughout the entire range of motion of the shoulder, reducing muscle activation by up to 53%. The pressure regulation system is also susceptible to outside disturbance, but can be remedied through the implementation of an adaptive control system that is capable of extreme disturbance rejection. Lastly, system identification can be performed in order to construct an efficient control system.

ACKNOWLEDGMENT

The authors would like to thank Ms. Tiana Miller-Jackson for her assistance during the preparation of the video and Ms. Marie Angela Ordonez for helping prepare the manuscript.

TABLE 2. RELATIVE CHANGE IN EMG RMS

| Movement | Target | Relative Reduction |
| --- | --- | --- |
| Abduction | L | 28.79% |
| Adduction | L | 33.89% |
| Horizontal Flexion | PM | 53.55% |
| Horizontal Extension | P | 25.87% |
| Forward Flexion | A | 39.73% |
| Forward Extension | P | 52.02% |

L = Lateral Deltoid, PM = Pectoralis Major, A = Anterior Deltoid, P = Posterior Deltoid